# Path Planning Considering Time-Varying and Uncertain Movement Speed in Multi-Robot Automatic Warehouses: Problem Formulation and Algorithm

**Jingchuan Chen[1], Wei Chen[2], Jing Li[2], Xiguang Wei[2], Wenzhe Tan[2], Zuo-Jun Max Shen[3,4,5], Hongbo Li[2]**

[1]Department of Industrial and Manufacturing Systems Engineering, The University of Hong Kong, Hong Kong, China
[2]Geekplus Technology Co., Ltd., Beijing 100102, China
[3]Faculty of Engineering, The University of Hong Kong, Hong Kong, China
[4]Faculty of Business and Economics, The University of Hong Kong, Hong Kong, China
[5]College of Engineering, University of California Berkeley, Berkeley CA 94720, USA.
jingchuan.chen@connect.hku.hk, wei.chen@geekplus.com, jing.li@geekplus.com, xiwang.wei@geekplus.com, wenzhe.tan@geekplus.com, maxshen@hku.hk, jason.li@geekplus.com

**Abstract:** Path planning in the multi-robot system refers to calculating a set of actions for each robot, which will move each robot to its goal without conflicting with other robots. Lately, the research topic has received significant attention for its extensive applications, such as airport ground, drone swarms, and automatic warehouses. Despite these available research results, most of the existing investigations are concerned with the cases of robots with a fixed movement speed without considering uncertainty. Therefore, in this work, we study the problem of path-planning in the multi-robot automatic warehouse context, which considers the time-varying and uncertain robots' movement speed. Specifically, the path-planning module searches a path with as few conflicts as possible for a single agent by calculating traffic cost based on customarily distributed conflict probability and combining it with the classic A* algorithm. However, this probability-based method cannot eliminate all conflicts, and speed's uncertainty will constantly cause new conflicts. As a supplement, we propose the other two modules. The conflict detection and re-planning module chooses objects requiring re-planning paths from the agents involved in different types of conflicts periodically by our designed rules. Also, at each step, the scheduling module fills up the agent's preserved queue and decides who has a higher priority when the same element is assigned to two agents simultaneously. Finally, we compare the proposed algorithm with other algorithms from academia and industry, and the results show that the proposed method is validated as the best performance.

## 1 Introduction

Path planning in the multi-robot system refers to calculate a set of actions for each robot, which will move each of the robots to its goal without conflicting with other robots. Lately, the research topic is received a great deal of attention for its extensive applications, such as airport ground [1], drone swarms [2], and automatic warehouses [3]. Despite these available research results, most of the existing investigations are concerned with the cases of robots with a fixed movement speed without considering uncertainty. In fact, there exist a variety of uncertainties in the real automatic warehouse, for instance, imprecise self-location, imperfect modeling of the surroundings, and the uncertain movements of other robots in the shared workplace. These uncertainties are manifested as the time-varying and uncertainty of speed in path planning. Therefore, in this work, we focus on the path planning for multi-robot with time-varying movement speed in the context of automatic warehouses.

Early studies on path planning focus only on single robot cases. The problem can be analyzed in a great number of ways, for instance, linear programming [4], satisfiability [5], and answer set programming [6]. Next, graph search algorithms (e.g., A* [7] and Jump Point Search [8]) are employed to calculate the minimal cost paths between the start vertex and the destination vertex on the graph. The algorithms of single-robot path planning provide a foundation for the analysis of path planning in multi-robot systems.

The issue of planning collision-free paths for more than one robot presents another layer of difficulty. In a general way, analyzing jointly for multi-robot requires calculating in a state space with the dimension increasing linearly with the number of robots. This means that the quantity of the state space is exponential in the number of robots. The literature of this direction can be divided into two groups for review, i.e., centralized and decentralized algorithms. For the centralized algorithm, a global search over the joint-space of all robots' movements is performed, and it outputs a joint-plan containing all robots' paths from their initial positions to their goal positions. Several representative centralized planning algorithms are summarized as follows: Paper [9] studies implementation of dimensional expansion for robot configuration spaces which can be denoted as a graph. Also, the conflict-based search algorithm is presented in [10]. The algorithm consists of two levels. At the high level, the search is carried out by a conflict tree based on conflicts between individual robots. At the low level, searches for single robot are performed. These algorithms provide guarantees of optimality and completeness but are considered as time-consuming. In the real-world multi-robot automatic warehouse, rather than spending significant amounts of resources calculating for optimal paths, it is instead preferable to plan valid, collision-free paths quickly, even if sub-optimal, and given additional time, to iteratively refine the paths.

As expected, distributing the computational cost among the robots can help to reduce the problem's computational complexity. Indeed, the state number of a decentralized algorithm can be made independent of its scale in principle [11]. Furthermore, decentralized algorithms are more robust to uncertainties and failures, compared to centralized algorithms. Therefore, they can be employed in the scenario where centralized algorithms are difficult to be used. For the related work of decentralized methods for path planning, some papers formulate the problem as reactive control or local coordination problems (see, for instance, [12], [13]). These algorithms scale with the size of robots but lack optimality and completeness guarantee. For the recent literature of decentralized methods, in [14], a decentralized multi-robot path-finding algorithm that can provide theoretical completeness and optimality guarantees is offered. a lifelong multi-agent path-finding problem is studied in [15], where robots are assigned with new goal locations constantly. The work considers the certainty caused by engaging the new task to robots constantly. Besides, paper [16] presents a multi-agent planning framework for a class of anytime planners that quickly generate valid paths with suboptimality estimates and generate optimal paths given sufficient time. However, unfortunately, these algorithms usually assume that robots can only move to a neighboring location or wait at their current location for each time step, and do not consider the uncertainty and time-dependent speed. Therefore, these algorithms cannot offer a guarantee to be implemented in real multi-robot environments.

We propose a decentralized algorithm considering time-vary and uncertain movement time in the automatic warehouse context to solve the above problem concerning path planning in multi-robot automatic warehouses. The major contribution of this work can be summarized as follows: 1) We formulate a real problem from industry, i.e., path finding for multi-robot with time-vary and uncertain movement speed in automatic warehouses context, into a mathematical model. 2) We propose an effective algorithm for path planning in a dynamic and uncertain surrounding, which consists of path-finding module, conflict detection and re-planning module, and scheduling module. 3) Under the framework of A* algorithm, we proposed the path-finding module, where the conflicts between robots is considered by using a normal distribution function to map the cost value. Under such arrangement, a great number of conflicts can be avoided when planning the path for each robot. 4) We design a conflict detection and re-planning module to handle the conflict existing in the planned path by re-plan robots' path. Besides, the scheduling module is used to settle part of conflicts by controlling the robots' movement.

The rest of the paper is organized as follows: The studied problem is formalized in Section 2. Then, in Section 3, we present the proposed algorithm. Section 4 demonstrates the results of the proposed algorithm in comparison to other algorithms. Finally, Section 5 sums up the paper and provides future work.

## 2 Problem formulation

### 2.1 Symbol definition

As shown in Fig. 1, considering a path planning problem in a grid world with grid $(x, y)$, we have a directed graph $G = (V, E)$.

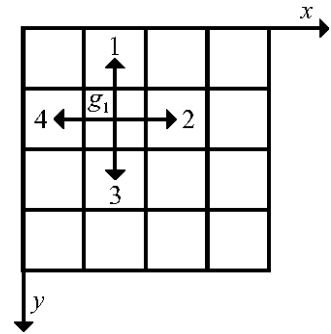

**Figure 1** Definition of robot's direction

Vertex set $V$ consists of grids and edge set $E$ describes the connectivity (relation of connection between vertexes). If vertex $V_i$ and $V_j$ connect directly, we have $e(i,j) \in E$. Remind that every grid connects with itself, so we also have $e(i,i) \in E$.

There are $m$ robots (denote as robot $r_i$, $i = 1,2,...,m$) in the automatic warehouse. For robot $r_i$, in time step $t$, a direction variable $d_i(t)$, $d_i(t) \in \{1,2,3,4\}$, is provided. The meaning of the value is showed in Fig. 1. We define the relation between the direction and the action grids as follows:

We assume the coordinate of current position $g_1 \in V$ is $(x_1, y_1)$ and the coordinate of next position $g_2 \in V$ is $(x_2, y_2)$. If $x_1 = x_2$ and $y_2 - y_1 = -1$, we have $g_2 - g_1 = 1$. If $x_1 = x_2$ and $y_2 - y_2 = 1$, we have $g_2 - g_1 = 3$. If $x_2 - x_1 = 1$ and $y_2 = y_1$, we have $g_2 - g_1 = 2$. If $x_2 - x_1 = -1$ and $y_1 = y_2$, we have $g_2 - g_1 = 4$.

A preserved queue $q_i(t)$, $q_i(t) \in V^N$, is also provided to show the grids obtained right now and the grids to be occupied in the future by the robot $r_i$. The maximum length of $q_i(t)$ is $N$, $N > 2$. The blank space in $q_i(t)$ is filled with placeholder #. The efficient length of preserved queue is $n_i \leq N$, ignoring the placeholder #. Robot has a phase variable $ph_i(t) \in R$, $ph_i(t) \geq 0$, to show the movement procedure from current position to the next. The phase variable is initialized to 0 when the movement starts. It is accumulated in every timestamp until it reaches 1 or more than 1, which means the movement has finished. When movement finishes, the first element in $q_i(t)$ will be removed to show that the occupation of this grid has finished. When $n_i \geq 1$, the robot will move strictly under the order of the grids in $q_i(t)$. When $n_i = 1$, the robot will stay. When the robot wants to change direction, it needs to stay for $W$ timestamps to change the direction and then move towards the next grid. Variable $w_i(t) \in Z$, $0 \leq w_i(t) \leq W$ is used to show the remained timestamps that the robot needs to stay. Every robot has a start position $S_i \in V$, a start direction $DS_i \in \{1,2,3,4\}$, a target position $G_i \in V$ and target direction $DG_i \in \{1,2,3,4\}$.

## 2.2 State space

In time step $t$, state $s_i(t) = [d_i(t), q_i(t), ph_i(t), w_i(t), (G_i, DG_i)]$ of robot $r_i$ consists of five parts: the direction of robot $d_i(t) \in \{1,2,3,4\}$, the occupied grid set $q_i(t) \in V^N$, the moving procedure set $ph_i(t) \in R$, $ph_i(t) \geq 0$, the variable that means the remained timestamps for robot to stay $w_i(t) \in Z$, $0 \leq w_i(t) \leq W$, the grid of target position and its direction $(G_i, DG_i)$, $G_i \in V$, $DG_i \in \{1,2,3,4\}$. In time step $t$, the united state space of $m$ robots $s(t) = [d(t), q(t), ph(t), w(t), (G, DG)]$ also consists of five parts: the direction of $m$ robots $d(t) \in \{1,2,3,4\}^m$ the occupied grid set of $m$ robots $q_i(t) \in V^{N \times m}$, the moving procedure of $m$ robots $ph(t) \in R^m$, $ph_i(t) \geq 0$, $i = 1,2,...,m$, the variables that means the remained timestamps for $m$ robots to stay $w(t) \in Z^m$, $0 \leq w_i(t) \leq W$, $i = 1,2,...,m$, the grid set of target positions of $m$ robots $(G, DG)$, $G \in V^m$, $DG \in \{1,2,3,4\}^m$.

## 2.3 Action space

In time step $t$, as long as the number of actual elements $n_i$ (the efficient length) in preserved queue $q_i(t)$ of robot $r_i$ is less than $N$, the algorithm could choose to append $\tilde{n}_i$ elements to the $q_i(t)$. We define action $a_i(t) \in V^N$ as the queue $q_i(t)$ of robot $r_i$ in time step $t$ which is after being appended (The blanks in $q_i(t)$ are filled in with the placeholder #).

## 2.4 State transition

The state transition describes the behavior of simulation environment. The united state $s(t + 1)$ in time step $t + 1$ is determined by the united state $s(t)$ and united action $a(t)$ in time step $t$. Robot $r_i$ needs to initialize state $s_i(0)$ when moving at the start position. We assume the robot has the direction $d_i(0) = DS_i$. The first element $q_i(0)$ in preserved queue is $S_i$ with placeholder # filling in other blanks, in other words,

$q_i^1(0) = S_i$, $q_i^l(0) = \#$, $l = 2,3,...,N$. The moving procedure phase and timestamp for staying are initialized to 0, $ph_i(0) = 0$, $w_i(0) = 0$. In every timestamp, we decide whether the robot needs to change its direction. If there is only one actual element or the robot direction $d_i(t)$ is the same with the action direction $[a_i^2(t) - a_i^1(t)]$ in preserved queue, then the robot will not change its direction. Otherwise, the robot needs to change its direction to the action direction and set the timestamp for staying $w_i(t) = W$.

After performing operation according to the direction, if $w_i(t) > 0$, the robot needs to stay. Robot has $q_i(t + 1) = a_i(t)$, $ph_i(t + 1) = 0$, $w_i(t + 1) = w_i(t) - 1$. If $w_i(t) = 0$, then moving procedure phase is update at first, $ph_i(t + 1) = ph_i(t) + v_i$, $v_i = \frac{f_i - 1}{N - 1} \times v$. $v$ is a fixed velocity given by the simulation system. $f_i$ means the number of elements before the first turning position in action, including the turning position itself. Remind that when there is only one actual element in action, $f_i = 1$. If $f_i = 1$, it means there is only one actual element in preserved queue. Agent will stay because of $v_i = 0$; If $f_i > 1$, $v_i > 0$. Agent will move. If $ph_i(t + 1) \geq 1$, it means action has finished. Then the first element in preserved queue needs to be deleted. If $ph_i(t + 1) < 1$, agent needs to stay until $ph_i$ is accumulated to no less than 1. In conclusion, we enumerate the transition steps of agent $r_i$ in time step $t$ as follows:

1) If $n_i + \tilde{n}_i = 1$ or $n_i + \tilde{n}_i > 1$, $d_i(t) = a_i^2(t) - a_i^1(t)$, we have $d_i(t + 1) = d_i(t)$. Otherwise, the robot needs to change direction and wait/stay. We have $d_i(t + 1) = a_i^2(t) - a_i^1(t)$, $w_i(t) = W$.

2) a) If $w_i(t) > 0$, we have $q_i(t + 1) = a_i(t)$, $ph_i(t + 1) = 0$, $w(t + 1) = w_i(t) - 1$. b) If $w_i(t) = 0$, we update moving procedure phase at first, $ph_i(t + 1) = ph_i(t) + v_i$, $w_i(t + 1) = w_i(t) - 1$, $v_i = \frac{f_i - 1}{N - 1} \times v$. Then, if $ph_i(t + 1) < 1$, we have $q_i(t + 1) = a_i(t)$, $w_i(t + 1) = 0$. If $ph_i(t + 1) \geq 1$, we have $q_i^l(t + 1) = a_i^{l+1}(t + 1) = a_i^{l+1}(t)$, $l = 1,2,...,n_i + \tilde{n}_i - 1$, $ph_i(t + 1) = 0$, $w_i(t + 1) = 0$.

## 2.5 Problem description

We assume the time for all robots from their start positions to their target positions, i.e., make-span, is $T$. The target of the algorithm is to find a path dynamically which consumes the least time for every robot as well as avoiding any kind of collision.

$$\min T \quad (1)$$

s.t. $a_i^l(t) = q_i^l(t), 1 \leq l \leq n_i, i = 1,2,...,m$ (2)

$a_i^l(t) \neq q_j^k(t), 1 \leq l \leq n_i + \tilde{n}_i,$

$\quad 1 \leq k \leq n_i + \tilde{n}_i, i \neq j, i,j = 1,2,...,m$ (3)

$a_i^l(t) \neq q_i^k(t), 1 \leq l < k < n_i + \tilde{n}_i, i = 1,2,...,m$ (4)

$e[a_i^l(t), a_i^{l+1}(t)] \in E, 1 \leq l \leq n_i + \tilde{n}_i - 1,$

$$i = 1,2,\ldots,m \quad (5)$$

$$n_i + \tilde{n}_i \leq N, i = 1,2,\ldots,m \quad (6)$$

where Equation (2) means elements in the preserved queue must exist in the first part of the action vector. Equation (3) and (4) mean the elements in all action tensors should not be shown up in more than one path, except the placeholder #. Equation (5) means the elements in each action vector must constitute a path. Equation (6) means the sum of appended elements and existed elements should be no more than the length of the preserved queue.

## 3 Algorithm

In this section, we provide an algorithm to solve the problem formulated in Section 2. It consists of three modules, i.e., path-finding module, conflict detection and re-planning module, and scheduling module. Specifically, the path-planning module searches a path with as few conflicts as possible for a single agent by calculating traffic cost based on normally distributed conflict probability and combining it with the classic A* algorithm. However, this probability-based method cannot eliminate all conflicts and the uncertainty of speed will cause new conflicts constantly. As a supplement, the other two modules are proposed. The conflict detection and re-planning module chooses objects that requiring re-planning paths from the agents involved in different types of conflicts periodically by our designed rules. Also at each step, the scheduling module fills up the agent's preserved queue and decides who has a higher priority when the same element is assigned to two agents simultaneously.

### 3.1 Path-finding module

Under the framework of the A* algorithm, we propose the following method to find robots' paths. Compared with the A* algorithm, the major difference of the proposed method is the calculation of cost function.

Intuitively, the conflict may occur when two robots arrive at the same position in a close time step, and potential conflicts can be predicted by analyzing the arrival time of each robot at each node. However, it is very difficult to estimate the arrival time for each robot because its speed is uncertain and its wait time alongside the path is hard to predict since our method is decentralized. Hence, in this work, the distance from a robot to a node is employed to approximate the elapsed time via the route. Let $dis_i$ and $dis_j$ represent the distance between the current location of robot $r_i$ and robot $r_j$ to the conflict node, respectively. The closer $|dis_i - dis_j|$ is to 0, the closer conflict probability is to 1, and conflicting probability will gradually decrease as $|dis_i - dis_j|$ increase. Considering this fact, we adopt the normal distribution function to map the cost function, and the calculation equation of cost function is offered as follows:

$$f(V) = g(V) + h(V) + t_{\text{traf}}(V) \quad (7)$$

where, $g(V)$ represents the actual cost of an optimal path from $q_i(t)$ to $V$ and $h(V)$ denotes the actual cost of an optimal path from $V$ to a preferred goal node of $V$ (see [7]). Besides, the calculation method of $t_{\text{traf}}(V)$ is offered as follows:

$$t_{\text{traf}}(V) = \sum_{i=1}^{3}\left[\sum_{j=1}^{m_i} \zeta_i e^{-\frac{\left(\frac{s-d_j}{\sigma}\right)^2}{2}} c_1^{-\frac{s+d_i}{2}}\right] c_2^{m_i} + c_3\beta \quad (8)$$

where, $m_i$, $i = 1,2,3$, refers to the number of opposite conflicts, following conflicts, and crossing conflicts occurring in the planned path, respectively. $\zeta_i$ denotes the base cost for different type of conflicts, $s$ is the distance between the current node and $V$, $d_j$ represents the distance between the current node and to the conflict node, $\beta$, $\beta \in \{0,1\}$, denotes the robot undergo a turn ($\beta = 1$) or not ($\beta = 0$) in the last time step, $c_1$, $c_2$, and $c_3$ denote the constants.

It should be noted that the definition and calculation method of the conflicts, i.e., calculating $m_i$, is provided in Section 3.2.

### 3.2 Conflict detection and re-planning module

The primary goal of path-finding solvers is to find paths that can be executed without collisions. To achieve this, we need to detect the potential conflict between robots and settle them by re-planning conflicting paths. First, we list the definition and judgment method of considered conflicts as follows:

- Opposite Conflict. There is an opposite conflict between robot $r_i$ and robot $r_j$ if and only if both the robots are planned to occupy the same vertex at the same time step or traverse the same edge in opposite direction.

- Following Conflict. There is a following conflict between robot $r_i$ and robot $r_j$ if and only if they occupy the same vertex in the same direction.

- Crossing Conflict. There is a crossing conflict between robot $r_i$ and robot $r_j$ when both the robots occupy the same vertex with their movement directions perpendicular to each other.

Fig. 2 illustrates the conflicts defined above, where the circles denote the robots, and the arrows represent robots' movement direction.

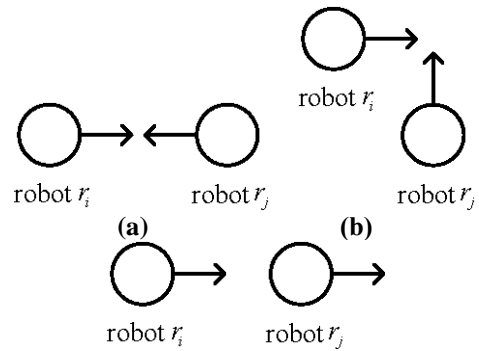

**(c)**

**Figure 2** An illustration of considered conflicts **(a)** opposite conflict **(b)** following conflict **(c)** crossing conflict

We define four user-specified parameters, namely time horizon $\tau$, threshold value for settling conflicts $\phi$, following conflict weight $\delta_{\text{fol}}$, and crossing conflict weight $\delta_{\text{cross}}$. Next, according to the conflicts defined above, we provide the following steps to determine the robots needed to be re-planed.

Step 1. According to the judgment method offered above, the algorithm detects all the conflicts between robots in the system within a time horizon of $\tau$ time steps.

Step 2. The algorithm re-plans the robot's path with the maximum number of opposite conflicts, and the procedure is repeated until there do not exist opposite conflicts in the planned paths for all robots.

Step 3. We calculate weighted conflict value $\gamma_i$, $i = 1,2,\ldots,m$, for all robots in the system as follows:

$$\gamma_i = \delta_{\text{fol}} n_{\text{fol},i} + \delta_{\text{cross}} n_{\text{cross},i} \qquad (9)$$

where, $n_{\text{fol},i}$ and $n_{\text{cross},i}$ refer to the number of following conflicts and crossing conflicts existing in the robot $r_i$'s path, respectively.

Then, the weighted conflict value of the system can be calculated by

$$\gamma = \max(\gamma_1, \gamma_2, \ldots, \gamma_m). \qquad (10)$$

Step 4. If $\gamma > \phi$, go to Step 5. Otherwise, the procedure ends.

Step 5. We re-plan the robot's path with maximum $\gamma_i$, and return to Step 3.

It should be noted that the algorithm treats the robot that has reached the target position as a static obstacle. Under such an arrangement, the conflict between robots and a robot having reached the target position need not be detected.

### 3.3 Scheduling model

According to the action space described in Section 2.3, the algorithm needs to append elements to the preserved queue when $n_i < N$. However, some conflicts may arise when more than one robot append the same node. To resolve the conflicts, we offer the following scheduling module, which offers a strategy to determine which robot is added the conflicting node to its queue.

Then, we assume that both robot $r_i$ and robot $r_j$ append the same node, i.e., $q_i^k(t) = q_j^g(t)$. We summarize the framework of the scheduling module as Algorithm 1. By using the algorithm, we can resolve the conflict by assigning the conflicting node to either of the robots. Two rules are used to schedule the robots when the algorithm does not effective (see line 19 of Algorithm 1).

Then, we offer three examples to further explain the scheduling module, as one can see from Fig. 3, where the star represents the target position of robot $r_j$.

**Algorithm 1** Scheduling Module
1: **if** $q_i^k(t) = G_i$ **then**
2:   **if** $q_i^{k-1}(t) = q_j^{g+1}(t)$ **then**
3:     **return** assigning the conflicting node to robot $r_i$'s preserved queue
4:   **else**
5:     **return** assigning the conflicting node to robot $r_j$'s preserved queue
6:   **end if**
7: **else if** $q_j^g(t) = G_j$ **then**
8:   **if** $q_i^{k+1}(t) = q_j^{g-1}(t)$ **then**
9:     **return** assigning the conflicting node to robot $r_j$'s preserved queue
10:   **else**
11:     **return** assigning the conflicting node to robot $r_i$'s preserved queue
12:   **end if**
13: **else**
14:   **if** $q_i^{k+1}(t) = q_j^{g-1}(t)$ and $q_i^{k-1} \neq q_j^{g+1}(t)$ **then**
15:     **return** assigning the conflicting node to robot $r_j$'s preserved queue
16:   **else if** $q_i^{k+1}(t) \neq q_j^{g-1}(t)$ and $q_i^{k-1} = q_j^{g+1}(t)$ **then**
17:     **return** assigning the conflicting node to robot $r_i$'s preserved queue
18:   **else**
19:     **return** assigning the conflicting node to the robot with a shorter distance to its destination or assigning randomly (if both the robots have a same distance to their destinations).
20:   **end if**
21: **end if**

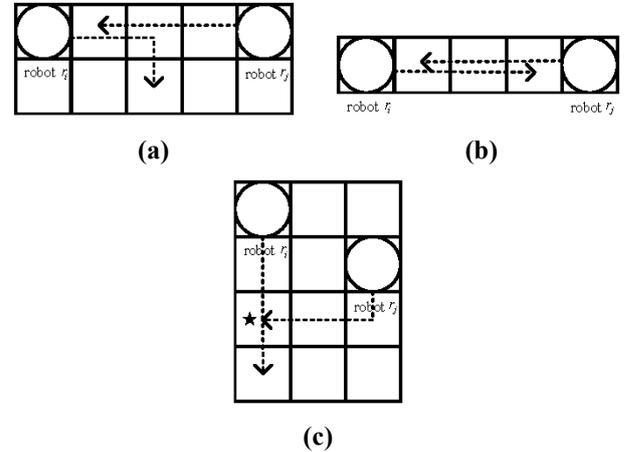

**Figure 3** Examples for explaining the scheduling module. **(a)** Case I **(b)** Case II **(c)** Case III

In case I, considering that $q_i^g \neq G_i$, $q_j^g \neq G_j$, $q_i^{k+1}(t) = q_j^{g-1}(t)$, and $q_i^{k-1} \neq q_j^{g+1}(t)$, the algorithm assigns the conflicting node to robot $r_j$. Moreover, in case II, we have that $q_i^k(t) \neq G_i$, $q_j^g(t) \neq G_j$, $q_i^{k+1}(t) \neq q_j^{g-1}(t)$, and $q_i^{k-1} \neq q_j^{g+1}(t)$. In this extreme case, the algorithm does not effective to assign the conflicting node, and two rules are adopted to settle the problem. Finally, for case III, it can be found that $q_j^g(t) = G_j$ and $q_i^{k+1}(t) \neq q_j^{g-1}(t)$. Clear, according to the algorithm assigns the conflicting node to robot $r_i$.

***Remark 1:*** Parameter $c_1$ and $c_2$ are the two major parameters in our algorithm, and they can be adjusted according to the size of the directed graph and the number of agents, respectively. The other parameters, i.e., $\xi_1$, $\xi_2$, $\xi_3$, $\sigma$, $c_3$, $N$, $\delta_{\text{fol}}$, $\delta_{\text{cross}}$, $\tau$, and $\phi$, are majorly chosen based on the hardware property of robots and they can be fixed in a certain warehouse.

## 4  Experiments

To demonstrate the superiority of the proposed algorithm, we carried out some comparison experiments. Specifically, we use a $30 \times 30$ grid world with 80 robots to simulate the multi-robot automatic warehouses as shown in Fig. 4, where the circle with a number denotes the robot and the square with a number represents the destination of the corresponding robot.

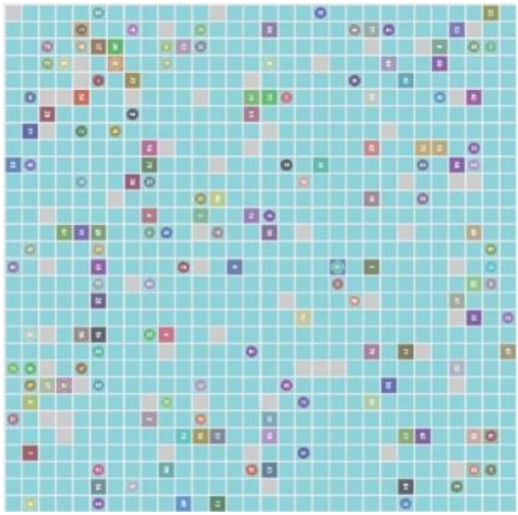

**Figure 4** Multi-robot automatic warehouse simulation environments

Quantitative comparisons with another three algorithms were carried out, and the description of these algorithms for comparison is offered as follows:

1) Proposed Algorithm with Different Cost Calculation (ADCC). Its framework is as same as the proposed algorithm. The only difference is the cost calculation method, which is provided as follows:

$$f_{\text{cost}} = c_1 \times \frac{n}{n_{\max}} \qquad (11)$$

where $c_1$ signified a constant, $n$ denotes how many times node $n$ will be visited by all robots, and $n_{\max}$ represents the maximum visit times among all nodes.

2) Cooperative A* (CA*) [17]-Based Algorithm. The algorithm is based on a prioritized-based polity. Specificity, within a time horizon of $\tau$ time step, each robot is randomly assigned a priority and computes, in priority order, the shortest path that does not collide with the paths of robots with higher priorities. In the algorithm, the path-finding process is divided into several single-agent searches, which are performed in three-dimension space-time. It considers the planned routes of other agents. A waiting move is covered in one agent's route to make it stay stationary. The states along the route are put into a reservation table after each agent's path is obtained. Entries in the reservation table are considered impassable and are avoided during searching by subsequent agents.

3) Priority-Based Search (PBS) [18]-Based Algorithm. The high level of the algorithm is similar to conflict-based search. The major differences are that the priority of node is represented by an adjacency matrix, and the child node inherits the priority of the parent node. Besides, the low level of the algorithm is similar to the cooperative A* algorithm. It carries a depth-first search on the high level to dynamically construct a priority ordering and thus sets a priority tree. Specifically, when occurs a collision, the algorithm chooses which agent can be offered a higher priority. It backtracks and explores other branches if and only if no solution is in the current branch. Therefore, it built a single partial priority ordering until it finds no collisions in the calculated path.

In addition, we offer the parameter section of the proposed method for the experiment as follows:

$$\xi_1 = 4, \quad \xi_2 = 1, \quad \xi_3 = 2, \quad \sigma = 4,$$
$$c_1 = 1.05, \quad c_2 = 1.5, \quad c_3 = 2, \quad N = 4,$$
$$\delta_{\text{fol}} = 1, \quad \delta_{\text{cross}} = 2, \quad \tau = 12, \quad \phi = 3.$$

Then, we evaluated the performance of these algorithms in terms of make-span. Considering the uncertainty of these algorithms, for each algorithm, we carried out 15 times for different $v$. The experiment results are summarized as box plots as shown in Fig. 5, and the average value of each case is offered in Table I, where "PA", "ADCC", "CA*", and "PBS" denote the proposed algorithm, proposed algorithm with different cost calculation, cooperative A*-based algorithm, and priority-based search-based algorithm, respectively.

**Table II** Average value of make-span for different cases

|  | $v = 1$ | $v \in [0.5, 1]$ | $v \in [0, 1]$ | $v = 0.5$ | $v \in [0, 0.5]$ |
|---|---|---|---|---|---|
| **PA** | 135.8 | 194.6 | 265.8 | 235.1 | 454.6 |
| **ADCC** | 147.4 | 205.2 | 270.8 | 241.1 | 472.3 |
| **CA*** | 287.1 | 461.7 | 691.7 | 644.3 | 1244.3 |
| **PBS** | 282.0 | 458.8 | 692.5 | 649.7 | 1288.4 |

As one can see from the figures and table, the proposed algorithm receives the best performance. Speed variation is a major concern of robot dynamics in our warehouse. Robots under different states can have different speeds (loaded robots are of max speed 0.5 and unloaded robots are of max speed 1). Therefore, we performed our experiment under different speed setting. As is shown in Fig. 5, the make-span of our proposed algorithm outperforms other algorithms under all speed settings and the variation of its make-span stays at a fixed range.

As one can see from Table 1, in terms of make-span, the proposed algorithm receives the best performance for all the cases. By comparing the first row with the second row, we can see that the proposed cost calculation method can further improve the performance. At the same time, it can be seen from the first, third, and fourth row, the proposed path-finding framework shows great improvement, even if using the traditional cost function.

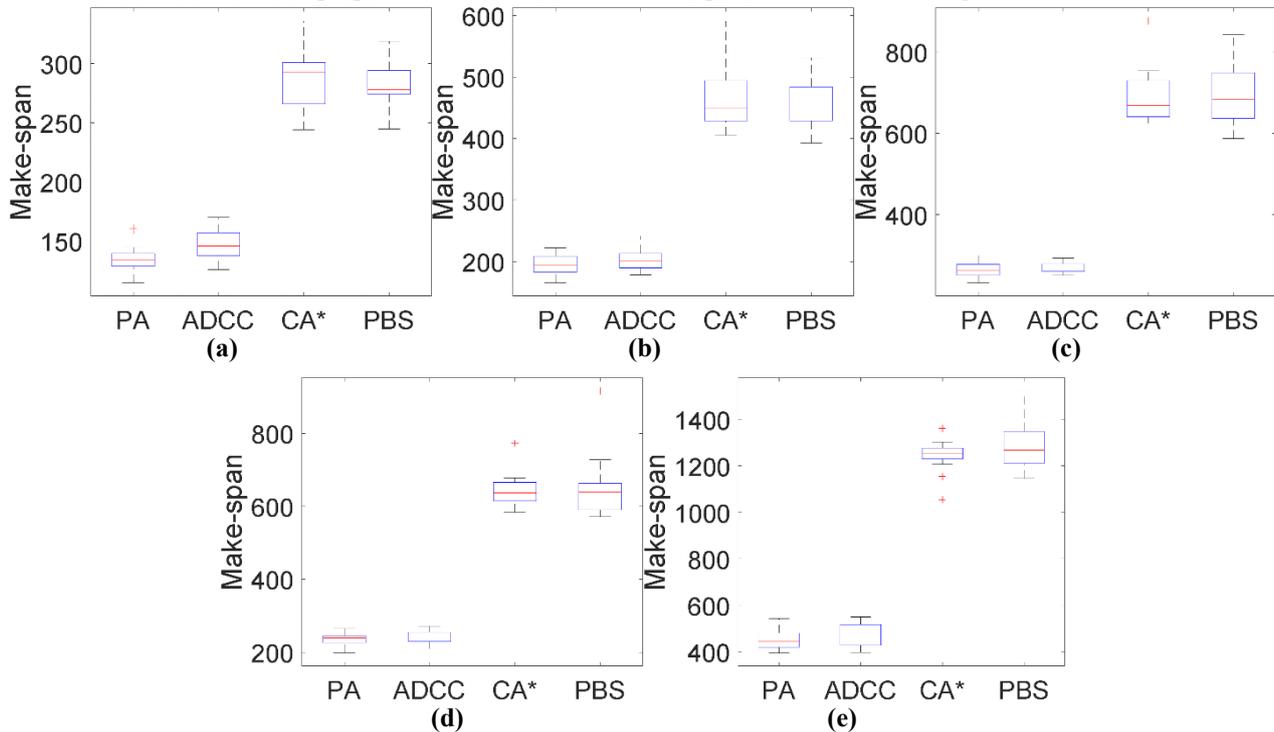

**Figure 5** Comparisons of the three algorithms in terms of make-span for different $v$ (a) $v = 1$ (b) $v \in [0.5,1]$ (c) $v \in [0,1]$ (d) $v = 0.5$ (e) $v \in [0,0.5]$

## 5  Conclusions and future work

In this work, we propose an algorithm for path planning in the context of automatic warehouses having multi-robot with time-vary and uncertain movement speed. Specifically, we first formulate the actual problem in multi-robot automatic warehouses. Then, we propose an algorithm to plan the paths, which consists of three modules, i.e., path-finding module, conflict detection and re-planning module, and scheduling module. The comparison experiments show that the proposed algorithm receives the best performance.

To further expand the study, we summarize the topics that can be investigated in the future as follows:

1) Employing some learning approaches (e.g., reinforcement learning and deep learning) to discover more advanced cooperative strategies and thereby obtain team-wide benefits with broader definitions.

2) Deploying incremental search techniques to reuse search effort from previous searches.

3) Exploring how to enable robots to plan their paths so that they do not have to frequently change their directions rapidly for avoiding obstacles for robots not good at quick turns.

4) Carrying out the proposed algorithm to a real automatic warehouse to further validate it.